# AI and Jobs: Has the Inflection Point Arrived?

# Evidence from an Online Labor Platform

**Dandan Qiao**, qiaodd@nus.edu.sg
National University of Singapore, 13 computing drive, Singapore, 117417

**Huaxia Rui**, huaxia.rui@simon.rochester.edu
University of Rochester, River Campus 500 Joseph C. Wilson Blvd., Rochester, New York, United States, 14627

**Qian Xiong**, xiongqian5@citic.com
CITIC Group Corporation, Beijing 100020, China

**Abstract**

This study investigates how artificial intelligence (AI) influences various online labor markets (OLMs) over time. Employing the Difference-in-Differences method, we discovered two distinct scenarios following ChatGPT's launch: displacement effects featuring reduced work volume and earnings, exemplified by translation & localization OLM; productivity effects featuring increased work volume and earnings, exemplified by web development OLM. To understand these opposite effects in a unified framework, we developed a Cournot competition model to identify an inflection point for each market. Before this point, human workers benefit from AI enhancements; beyond this point, human workers would be replaced. Further analyzing the progression from ChatGPT 3.5 to 4.0, we found three effect scenarios, reinforcing our inflection point conjecture. Heterogeneous analyses reveal that U.S. web developers tend to benefit more from ChatGPT's launch compared to their counterparts in other regions. Experienced translators seem more likely to exit the market than less experienced translators.

# Introduction

Thanks to the tremendous growth in computation power and data volume, artificial intelligence (AI) has advanced significantly over the past decade and started to permeate all walks of life. Among the most transformative advances is the rise of Large Language Models (LLMs) – AI systems with remarkable skills in simulating human-like abilities across a wide range of language-related tasks [11, 43]. What began with the debut of ChatGPT in late 2022 has now evolved into a global wave of LLM assimilation, where such models are embedded across industries, rapidly becoming an indispensable tool for many individuals and organizations [20, 38, 53]. For example, from college writing to bar exams, ChatGPT has repeatedly surprised people with its astonishing capabilities [39]. Many markets are exposed to this powerful AI tool, renewing the debate of the “technology displacement”, an issue extensively studied in macroeconomics and labor economics, especially during the 1990s in the wake of computerization across many industries [7, 37, 45]. Recent progress in agentic AI further adds urgency to this critical issue.

At the heart of the debate is the power of information technology (IT) to automate many tasks, thereby enhancing the productivity of human labor but also potentially leading to the substitution of labor by technology [2, 4, 8]. Following Acemoglu and Restrepo [2], we refer to these two opposing effects as the productivity effect and the displacement effect, respectively. These two effects jointly shape the effect of IT in general and AI in particular on human labor. In this fruitful literature, technology is treated as a black box, entering an economy’s production function as a factor alongside human labor in an aggregated manner. This macroscopic approach is taken by economists to study the long-term impact of a general automation technology. While this approach has yielded valuable insights into the historical effects of automation technologies, the rapid advancement of current wave of AI technologies, particularly LLMs, necessitates a more granular analysis. LLMs have diffused at an unprecedented scale and speed—tools like ChatGPT reached millions of users worldwide within months—producing labor market effects that diverge significantly from those associated with previous technological waves. This swift and widespread accessibility—both in terms of functional scope and individual usability—makes it imperative to understand the more immediate and micro-level impacts of AI, especially on individual workers across different markets.

In addition, prior literature often relies on a task-based framework to assess the impact of IT, categorizing tasks according to their susceptibility to automation by IT or AI systems [8, 2, 4]. These frameworks have yielded valuable insights, such as distinguishing between routine and non-routine tasks or separating prediction from decision-making. However, they are typically grounded in the dominant technical capabilities of a specific technological wave. As AI continues to evolve rapidly across multiple dimensions—ranging from language modeling to multimodal reasoning, and from assistive tools to autonomous agents [11], tasks that were once considered non-automatable may soon fall within the scope of AI systems. Thus, existing task-based frameworks may become obsolete as AI's intelligence frontier advances.

In response, our study introduces a micro-level, technology-agnostic framework, i.e., the inflection point conjecture. Rather than relying on fixed task typologies, our framework centers on the proportion of a job's tasks that can be successfully completed by AI. It considers both demand-side effects (e.g., shifts in market potential as clients adopt AI) and supply-side effects (e.g., changes in freelancers' cost structures and productivity). This framework offers a generalized lens to understand both productivity and displacement effects in the context of ongoing AI advancement across different labor markets.

As per the research goal, our study focuses on understanding the more immediate effects of AI on the labor market at the individual level, particularly within online labor markets (OLMs). Unlike full-time jobs that are more stable, freelance jobs are more susceptible to changes in market conditions [5, 26]. We expect the impact of major AI innovations on jobs to first unfold on freelance markets. Thus, to understand the labor market implications of the current wave of AI innovations, we study in this paper the impact of ChatGPT on workers on an online freelance platform. With this empirical context, we can take advantage of the micro-level data available there for empirical investigation. Indeed, a significant barrier to assessing the impact of AI on the workforce has been the absence of high-quality data, obstructing in-depth and timely empirical analysis at more granular levels [22]. Previous studies of the relation between IT/AI and labor usually focus on macro-level long-term industry dynamics, which, while valuable, may not adequately capture the immediate and nuanced impacts of AI on individual workers [2, 10, 51]. Although several recent studies have empirically examined the impact of ChatGPT on freelancers, their findings primarily highlight the displacement effects within one or two specific markets [16, 29, 34]. However, the question of whether advancement in AI substitutes or complements human workers cannot be answered in binary terms. How this wave of LLM innovations affects individual workers across different markets remains unclear. Our study will fill these gaps and provide a more comprehensive analysis of both the displacement and productivity effects that ChatGPT may have on freelancers.

Specifically, we collected data from one of the most popular freelance platforms, which provides a hierarchical freelancer classification system and accessibility to complete work records. We aggregated the information at the worker level on a monthly basis to compile a dataset spanning from May 1, 2022, to October 30, 2023. Through a Difference-in-Differences (DiD) design, we discovered two contrasting scenarios where ChatGPT impacts freelancers in two opposite directions: 1) the displacement effect for translation & localization OLM where freelancers' work volume and earnings decreased significantly after the release of ChatGPT; 2) the productivity effect for web development OLM where freelancers' work volume and earnings increased significantly after the release of ChatGPT. A series of robustness checks was also conducted to further test the validity of these findings.

To better understand the underlying economic mechanisms that drive the two contrasting scenarios, we developed a microeconomic model of freelancers based on Cournot competition, where AI reduces both the market potential due to its displacement effect and the marginal cost due to its productivity effect. Despite its simplicity, the model implies the existence of an inflection point for each market. Before AI performance reaches the inflection point, freelancers benefit from any progress in AI performance, but after crossing the inflection point, any further improvement in AI performance will hurt freelancers. Because the relative position of AI performance and the inflection point differs by OLM, this inflection point conjecture explains the two contrasting scenarios observed in the translation & localization OLM and the web development OLM.

To shed light on the generalizability of our empirical findings and further test the inflection point conjecture, we collected data from eleven additional OLMs and consider the release of ChatGPT 4.0 as another improvement of AI. Two important patterns emerge from this comprehensive empirical exercise. First, estimating both the effect of ChatGPT 3.5 and ChatGPT 4.0 on freelancers in all OLMs reveals three scenarios: 1) displacement effects in both AI advances; 2) productivity effects in both AI advances; and 3) productivity effect followed by displacement effect. The noticeable absence of a transition from a net displacement effect to a net productivity effect is in line with the inflection point conjecture which suggests that once the displacement effect dominates, it cannot be reversed. Second, our analysis reveals heterogeneous effects across occupational categories, which we classified into five clusters based on their exposure to AI. Writing jobs appear most vulnerable to displacement. Consulting and programming jobs initially benefit from productivity gains but may face future substitution as AI capabilities advance. Operational and creative jobs—where human judgment or originality is essential—mainly experience productivity-enhancing effects, at least for now.

We also did additional empirical analyses to further enrich our findings. The extended-timeframe analysis indicates that as AI's capability evolves, the displacement effects observed in translation & localization OLM tend to intensify, and web development OLM could finally reach its inflection point, switching from productivity effects to displacement effects gradually. An analysis based on the weekly fulfilled demand of each OLM confirms a decline in total transaction volume for OLMs where the displacement effect dominates, and an increase in total transaction volume for OLMs where the productivity effect dominates. Moreover, a worker-level heterogeneity analysis reveals that freelancer location has a moderating effect for the web development OLM but not on the translation & localization OLM, which is in line with our proposed mechanism. Freelancer location is a supply-side factor related to whether a freelancer can easily leverage ChatGPT for productivity enhancement, which, hence, may moderate the productivity effect but not the displacement effect.

The remainder of the paper proceeds as follows. We first review several streams of related literature. Next, we conduct empirical analyses on two representative markets—translation & localization and web development—revealing the displacement and productivity effects, respectively. We then develop the inflection point conjecture to explain how the interaction between these two effects generates opposite outcomes across markets, followed by additional analyses to test the proposed mechanism and assess its external validity. Subsequent analyses explore heterogeneity across freelancers. Finally, we summarize the study's contributions, limitations, and implications for future research.

# Research Background

## Impact of Automation Technology on Labor Market

In the past decades, automation technology has seen tremendous development, raising concerns in relation to "technological unemployment". To a large extent, automation technology eliminates the demand for labor undertaking repeated and manual work. Such displacement has shifted the labor demand towards skilled and highly educated ones [7]. However, at the same time, researchers have also acknowledged automation technology as an effective tool to augment human ability, enhancing their competence in the labor market [8, 24]. Some studies further demonstrated that these technologies have the potential to create new industries and job opportunities for human labor [1, 2]. These mixed effects (i.e., displacement and productivity effects) give rise to an important research branch exploring the relation between automation technology and labor.

Economists have engaged in extensive theoretical deliberation to understand how automation technology might impact human labor. Some research utilizes economic models to describe the elasticity of substitution among different production factors, such as IT, labor, and capital [18, 51]. Other research has extensively explored the role of technology in working processes [7, 8]. Notably, Autor et al. [8] introduced the perspective of task composition to explain how computer technology alters tasks within a market and subsequently affects the demand for human skills. Specifically, routine tasks, governed by explicit rules, are readily automated, whereas nonroutine tasks, lacking defined rules, primarily experience a productivity effect with automation technologies. This "Routine-biased Technological Change" perspective is widely acknowledged for understanding how technological change impacts various types of human labor.

With the advancement of AI, some scholars have tried to extend the theoretical model from this prior literature to understand the impact of AI [2, 3, 4]. For example, Acemoglu and Restrepo [2] employed a task-based approach to show that automation, specifically AI and robotics, extensively displaces human labor. Nonetheless, they also emphasized the presence of countervailing aspects with the potential to mitigate this displacement effect. Acknowledging AI's premier capability in prediction, Agrawal et al. [4] delineated jobs into prediction and decision tasks, suggesting that AI's impact on various labor markets could be ambiguous. While these studies shed light on the relationship between IT/AI and human labor, their investigations, typically conducted by macro- and labor economists, focus on the long-term effects of general automation technologies at a broad, macro level. They tend to overlook the detailed, immediate impact of specific technologies on individual workers at a micro level. Our study departs from this economic literature by offering a more granular analysis. We develop a formal inflection point framework grounded in economic modeling that explains why and when the impact of AI shifts from productivity to displacement, thus leading to heterogeneous impacts on individual freelancers observed across different markets. In addition, unlike the prior literature, which often takes a cross-sectional view with technology either complementing labor (in some markets or for some workers) or substituting labor (in some other markets or for some other workers), our theoretical model offers a temporal perspective that allows us to see how different effects unfold in sequence for each market. This perspective not only allows us to understand the opposing effects of the same technology at the same time in different markets, but also gives us a way to think about the long-term implications as AI relentlessly advances.

There are also some empirical attempts in recent decades to study the impact of automation technology on labor markets. However, these studies have yielded mixed results and remain inconclusive. At the aggregate level, while some found a net displacement effect, some found evidence for a net productivity effect [10, 15]. At the micro level, however, the impact often depends on different types of employers or workers [35, 52]. For instance, Lu et al. [35] showed that, in the context of health IT adoption, licensed nurse staffing levels increased in low-end nursing homes but decreased in high-end nursing homes. Zhang et al. [52] proved that while highly educated labor received a productivity effect and less educated labor received a displacement effect, the net effects on the averagely educated labor depended on task routineness.

When the focus shifts to AI, dual effects are also present in the labor market, aligning with findings observed in broader automation technology studies [36, 49, 52]. For instance, Lysyakov and Viswanathan [36] revealed that lower-tier designers tend to exit the online market when facing the threat of image-generating AI, while high-tier designers could become more engaged. Xue et al. [49] demonstrated that increasing AI applications positively impacts the employment of non-academically trained workers in firms, yet adversely affects academically trained employees, which collectively indicates a net positive effect on overall employment. However, these studies primarily rely on data from a single market or macro-level analysis, which cannot capture the varied effects of AI across different workers and labor markets. Recent research has also shown that generative AI triggers heterogeneous behavioral responses at the individual level [42]. Such data limits have become a significant barrier in comprehending the contextual impact of AI on the workforce [4].

Leveraging the recent advancement of large language models, our research adopts a multi-market approach to systematically assess how the same AI shock—the release of ChatGPT—has generated heterogeneous effects across a wide range of online labor markets. By analyzing two consecutive shocks (ChatGPT-3.5 and ChatGPT 4.0), we further examine how these effects evolve with advancements in AI capability. Moreover, our findings show that the impact of AI varies not only across job categories but also by freelancer experience and geographic location, offering a more nuanced and fine-grained understanding of AI-labor dynamics.

## Online Labor Market

The online labor market (OLM) has grown tremendously in the past decades. The OLM has shifted the traditional labor market onto online platforms, introducing new avenues for labor transactions in the digital economy. By joining an OLM, workers can access job opportunities beyond national boundaries, actively participating in the global labor market instead of being confined to local demand [31]. The emergence of this market also benefits employers by enabling platform-mediated transactions and communication, thereby reducing transaction costs [26]. By 2021, more than 160 million user accounts have been registered as online freelancers.[1] On one hand, the unique attributes of OLMs yield substantial social benefits, such as mitigating offline unemployment and enhancing the well-being of workers in developing countries [28, 31]. On the other hand, these digitalized attributes facilitate the inherent flexibility of worker mobility within OLMs and magnify the immediate and widespread impacts of AI [50, 5, 26]. The nature of short-term employment in OLMs further makes online freelancers particularly vulnerable to AI-induced market disruptions. Given the significant role of OLMs in the global labor market, comprehensively understanding the impact of AI on OLMs is crucial, which will be the focus of our study.

Existing literature on OLM can be categorized into three streams, corresponding to the focus on workers, employers, and the platform. From the perspective of labor supply, OLM is an alternative marketplace for employment and serves as an influential and effective offset for offline unemployment [28]. Researchers also focus on workers' well-being, highlighting the significant roles of reputation and skills in determining their market value [33]. From the perspective of labor demand, existing literature mainly tries to answer how an employer can optimize the hiring decision. A key factor is the employer's reputation, aiding in attracting superior talent and streamlining transaction and negotiation processes [9]. From the platform's standpoint, academic research primarily concentrates on fostering effective communication between online employers and workers as well as optimizing operations, such as strategies for platform incentives and bid auctions [25].

[1] Oxford Internet Institute: https://ilabour.oii.ox.ac.uk/how-many-online-workers/

OLM’s basis on AI-exposed digital platforms has magnified the extensive impact of automation technology [5, 26]. Meanwhile, AI agents increasingly operate as active teammates rather than passive tools [17]. This spurs a recent wave of literature dedicated to algorithm-based features to facilitate employee-employer matching from the perspective of platform operations [27, 32]. For instance, Horton [27] conducted a field experiment and demonstrated that algorithmic recommendations could significantly help employers fill their online technical job vacancies. Kokkodis and Ipeirotis [32] considered job-application characteristics to further improve the recommendation system for OLMs.

However, the existing literature on AI and OLMs primarily focuses on the platform operation [27, 32]. The extent to which LLMs affect freelancers across various OLMs remains insufficiently studied. By using ChatGPT’s release as an exogenous shock, this study aims to provide both empirical answers and theoretical explanations for how AI impacts freelancers across different markets.

## Large Language Models (LLMs)

Large Language Models have emerged as a revolutionary advancement in the realm of AI. The development of LLMs aims to address limitations in existing machine learning (ML) systems, which rely on supervised learning for language understanding [43]. These conventional ML systems typically function as supervised learners, which are trained from limited-domain datasets and are sensitive to data distributions, resulting in their lack of generalization. LLMs have freed themselves from reliance on explicit supervision and are instead pretrained on extensive general-purpose internet data to achieve the goal of maximally mimicking human language. In this pretraining process, LLMs naturally assimilate all relevant linguistic information and knowledge for language generation, which endows LLMs with innate abilities to process various downstream applications [11]. For instance, LLMs are frequently utilized for the efficient completion of tasks like translation and writing by analyzing the given prompts, as evidenced in prior work that highlights their use in assisting with ad copy creation [14]. This is known as “in-context learning” [46], which means that LLMs can adapt to diverse tasks without altering their internal structure, merely by integrating specific instructions or examples within their input.

Studies have attempted to both practically and theoretically explain the mechanisms behind the “in-context learnability” of LLMs [11, 43]. Despite being initially configured to maximize the probability of predicting unlabeled internet texts during pretraining, LLMs inherently acquire a wide array of abilities for language understanding and relevant task execution. Once these competencies are acquired and embedded through pretraining, “in-context learning” in LLMs primarily involves recognizing and applying these capabilities in response to specific instructional inputs for varied tasks [46]. This method closely mirrors the human approach to task processing, where understanding and action are derived directly from textual instructions.

The emergent abilities endowed by the pretraining process allow LLMs to contribute to various labor sectors. A notable instance is the release of ChatGPT, which brings the application of LLM to the general public and has rapidly become a valuable tool for individuals and organizations. Since its release, ChatGPT has reportedly amassed around 100 million active users monthly, setting a new record as the fastest-growing consumer app ever. Careers from different domains have been exposed to this popular AI tool [20], sparking the debate of AI displacing workers.

On the one hand, LLMs have the potential to act similarly to human labor by interpreting and executing tasks based solely on text-based instructions. As cost-effective and high-quality labor alternatives, LLMs might pose a significant challenge to the role of and even the necessity for human labor in certain markets [20]. On the other hand, the evolution of LLMs is leaning towards reducing barriers to entry into various markets by enhancing AI’s comprehension capabilities [46], potentially benefiting employees across diverse skill levels. Although numerous debates and discussions have taken place, there remains a lack of empirical investigation into the impact of ChatGPT on the labor market.

While several recent concurrent papers have also attempted to investigate the impact of ChatGPT's release on freelancers [16, 29, 34], their focus has primarily been limited to one or two specific job categories (e.g., writing). Moreover, these studies have only showcased the displacement effect that ChatGPT can have on these freelancers, namely, decreasing their transaction volumes or earnings. However, AI's impact on human workers is twofold: while it can enhance productivity, it can also reduce job opportunities. This issue should not be approached monotonically. Our study reveals a more complex relation between AI and jobs, both theoretically and empirically. In particular, we examine multiple OLMs to reveal both the displacement effect and the productivity effect of ChatGPT, and propose the inflection point conjecture to theoretically explain our empirical findings. We further explore the evolving role of AI in labor markets by leveraging the introduction of ChatGPT 3.5 and 4.0 as two natural shocks. This allows us to track the temporal dynamics of AI's impact and demonstrate how the same market may transition from productivity-enhancing to displacement-dominant over time.

# A Tale of Two Markets

## Empirical Context

Unlike full-time jobs that are more stable, freelance jobs are more susceptible to changes in market conditions. We expect the impact of major AI innovations on jobs to first unfold on freelance markets.[2] Hence, we undertake empirical analyses using data from a popular online freelance platform. This platform serves freelancers and clients across more than 180 countries, establishing a global labor market. It embraces the impact of AI on the labor market, permitting freelancers to utilize ChatGPT in their work. Jobs on this platform cover a large variety, such as translation, writing, web development, construction, and accounting, which allows us to examine how AI influences different OLMs. The jobs posted on this platform can be classified into two types depending on their price specification, i.e., fixed-price jobs and hourly-rated jobs. The fixed-priced job openings provide the total amount of compensation for the job, while the hourly-rated job openings provide a guide for the hourly price of the job and the estimated duration of the job. After a job is posted, workers who are interested can submit their proposals to the employer. Subsequently, the employer will review these applications and work proposals to select appropriate workers for the job vacancies.  Upon completion of the work, the employer releases the payment due and provides ratings and reviews for the worker based on the quality of the work. The platform has a hierarchical freelancer classification system that spans from a broad “category” to a narrower “subcategory” and more granular “specialties”. As shown in Figure 1, this platform categorizes all freelancers into 12 broad “categories”, each containing at least two “subcategories”, based on the jobs they have taken and the skills listed in their profiles. This platform also provides an advanced search feature that allows users to filter freelancers by category, subcategory, or specialty. This detailed system offers a clear portrayal of jobs necessitating specialized skills and corresponding human labor in OLMs, which allows us to obtain worker-level transaction histories related to distinct labor markets. Besides, the platform grants full access to the entire work history of its workers, including specifics such as job titles, received ratings, job start and end dates, job prices, and comments from employers. This enables us to accurately measure the acceptance time, completion time, and payment for jobs undertaken by workers since their registration. All recorded work histories represent deals that have been successfully transacted on the markets.

[2] We would like to clarify three key terms used throughout the paper, i.e., occupation, job, and task. Firstly, an occupation represents a category of jobs within a marketplace, which in the context of this study is often referred to as an OLM. Second, a job is a concrete project or work posted on the freelance platform. Lastly, a task is the smallest cognitive unit required for the successful completion of a job. By definition, a job consists of multiple tasks. Our empirical analyses and the economic model are based on the job, while the task is largely conceptual and implicit in this paper.

To meet our research objectives, we utilize ChatGPT’s release as an exogenous shock. Released on November 30, 2022, ChatGPT demonstrated high performance across various fields and became the first generative AI tool to gain mainstream recognition, making it an ideal candidate for studying the labor market implications of LLMs. Our initial analyses focus on two markets on the platform: translation & localization and web development, as LLMs have exhibited remarkable proficiency in performing relevant tasks. Construction design OLM was selected as the control group, given its lower susceptibility to automation by LLMs during the study period. Table A-1 in the Online Appendix A provides a summary of "specialties" belonging to different OLMs used in our initial analyses.

The capability of LLMs to manage a wide range of translation-related tasks has been thoroughly validated in real-world settings [41]. Researchers and practitioners have demonstrated ChatGPT's competitiveness against popular translation tools like Google Translate and its excellent ability to generate contextually relevant content [44]. Moreover, ChatGPT exhibits above-average performance in some language exams than human beings [39]. Therefore, we selected the translation & localization OLM as the quintessential market where the displacement effects of AI are expected to be more salient.

On the other hand, recent research has found that by using GitHub Copilot, a tool powered by OpenAI's generative AI model, web developers can implement an HTTP server in JavaScript 55.8% faster than developers without access to this AI tool [40]. Web development jobs involve a variety of tasks, including both front-end and back-end development, and require skills for both low-level implementation and high-level design. These multifaceted tasks might demand a comprehensive skill set, such as programming proficiency, problem-solving skills, debugging, systematic planning, and design expertise. Although ChatGPT cannot autonomously finish all these tasks, it has been demonstrated to play a supportive role to human programmers, assisting in tasks like code debugging and function identification. Therefore, we chose the web development OLM to explore the productivity effect of AI on freelancers.

Finally, we selected the construction design OLM as the comparison group, which has been demonstrated as one of the least impacted industries by ChatGPT [20]. ChatGPT's effectiveness in various tasks is largely driven by its ability to learn from large-scale internet-based training data. However, construction design tasks typically involve confidential, proprietary information and require domain-specific expertise. As a result, the availability of online relevant training data for these tasks is extremely limited. This data scarcity restricts ChatGPT's ability to generalize to construction design, making its influence on this market minimal during our study period. Researchers in the architecture, engineering, and construction (AEC) sector have also pointed out its slow rate of digitalization, due to its fragmented structure and reliance on specialized skills [47]. We also conducted several empirical analyses to validate the appropriateness of using construction design OLM as the control group. First, we utilized the AI Occupational Exposure (AIOE) Index, a metric developed to assess the extent to which various occupations are exposed to language model-based AI technologies [21]. Our analysis shows that construction design has a notably low AIOE score, indicating that it is relatively insulated from the influence of tools like ChatGPT.[3] Second, Google Search Volume Index (SVI) data shows that construction design exhibits a minimal (near-zero) Google SVI when searched alongside ChatGPT.[4] We also conducted two robustness checks to further confirm that our findings are not sensitive to the specific choice of control group. These findings collectively support the validity of using construction design as our comparison market. Therefore, freelancers on the construction design OLM, once appropriately matched, can serve as a good control group.

In the subsequent section, we mainly focus on these three markets to unveil the varied impacts that AI can bring to different OLMs. We later expand our study to include a broad spectrum of other OLMs for additional empirical investigations. Figure 2 provides an overview of all our empirical analyses on different OLMs of this platform, outlining the data sources, analysis unit, and primary objectives for each set of empirical analyses.

[3] The detailed results can be seen in the Figure B-1 of the Online Appendix B.
[4] SVI was obtained by querying the co-search frequency of "ChatGPT" and each market name using Google Trends. The plots are presented in Figure B-2 and Figure B-3 of the Online Appendix B.

## Data and Variables

To collect data for our empirical analysis, we identified workers engaging in each of the three aforementioned OLMs. We first determined the relevant "specialties" of these OLMs on the platform, based on their job content and skill requirements. Subsequently, we used the advanced search feature to identify the corresponding freelancers and obtain their work history data. In total, we obtained profiles and work histories of 6,293 unique workers belonging to the construction design OLM, 7,181 unique workers belonging to the translation & localization OLM, and 13,230 unique workers belonging to the web development OLM. We then removed those inactive workers who had not accepted any job before November 1, 2022, and aggregated the data at the worker level on a monthly basis. A worker within a specific market may possess multiple skills enabling them to engage in jobs beyond their primary OLM. In this paper, we define jobs aligned with workers' primary labor market as "focal jobs".

The goal of our empirical study is to analyze the impact of AI on freelancers; we hence focus on each worker's focal jobs within each OLM in the analysis. All measurements were constructed based on the focal jobs accepted within a given month, rather than those completed. We excluded data from November and December of 2022 to minimize the holiday effect and potential anticipation effect of pre-release activities. Hence, the study's time frame spans six months before the shock (i.e., May through October in 2022) and ten months after the shock (i.e., January through October in 2023). Table 1 provides the definitions of key variables, while Table 2 reports their descriptive statistics for each of the three OLMs in the main analyses.

## Identification Strategy

To examine the impact of AI on freelancers, we used the following two-way fixed-effect DiD model for identification, where the unit of analysis is at the worker-month level.

$$Y_{it} = \beta_0 + \beta_1 \times ChatGPT_{it} + \beta_2 \times X_{it} + \eta_i + \tau_t + \epsilon_{it} \quad (1)$$

In Equation (1), $i$ and $t$ index worker and month, respectively. The dependent variable $Y_{it}$ measures worker $i$'s transaction volume or total earnings in the focal OLM during month $t$. For the transaction volume, we use log($Fjobnum_{it}$) to measure the log-transformed number of focal jobs worker $i$ accepts in month $t$. For earnings, we use log($Fjobearn_{it}$) to measure worker $i$'s total earnings from focal jobs in month $t$. The explanatory variable of interest is the binary variable $ChatGPT_{it}$ (i.e., $Treat_i \times After_t$), which equals 1 if worker $i$ mainly belongs to the treated market and the transaction activities under investigation occurred after the release of ChatGPT. Otherwise, the binary variable $ChatGPT_{it}$ equals 0. $\eta_i$ captures the worker fixed effect, while $\tau_t$ captures the time fixed effect. $X_{it}$ captures all time-varying variables, such as workers' tenure measured by the number of months up to month $t$ since worker $i$'s registration. We clustered the standard error at the worker level.

To ensure workers in the treated and control groups are comparable, we used Propensity Score Matching (PSM) to improve the sample balance by accounting for workers' experience, total number of accepted focal jobs, wages (i.e., average price and hourly rate of focal jobs), and quality of work (i.e., the average rating of focal jobs). All these variables were calculated from the work record before ChatGPT's release. We adopted a 1:1 nearest-neighbor matching strategy at the worker level and excluded observations falling outside of the common support region [12].

## Effects on Translation & Localization Freelancers

Our first analysis aims to examine the effect of ChatGPT on translation workers, using comparable workers in the construction design OLM as the control group. After matching with a caliper value of $2 \times 10^{-4}$, we obtained 2,276 workers. Table 3 reports the balance test results before and after the matching. The results of the PSM-DiD estimation are presented in Table 4. We also performed a DiD analysis without matching, and the results are consistent with our main findings (see Table A-2 in the Online Appendix A).

Overall, we find strong displacement effects of ChatGPT on workers in the translation & localization OLM. More specifically, in column (1), which corresponds to the dependent variable of log($Fjobnum_{it}$), we find the coefficient of $ChatGPT_{it}$ negative and statistically significant, suggesting a decrease in the number of focal jobs accepted by workers after the release of ChatGPT. In terms of magnitude, the transaction volume dropped by 9.0% ($= 1 - e^{-0.094}$) on average. In column (2), which corresponds to the dependent variable log($Fjobearn_{it}$), the coefficient of $ChatGPT_{it}$ is also negative and statistically significant, suggesting a decrease in workers' earnings from focal jobs after the release of ChatGPT by 29.7% ($= 1 - e^{-0.353}$) on average.

The negative impacts of ChatGPT on the translation & localization market make sense. Pretrained on a vast amount of internet text, ChatGPT is particularly skilled at grammar, language comprehension, and translation. On the demand side, these strengths enable ChatGPT to deliver high-quality translation services. As a result, employers can now complete translation jobs efficiently and at a lower cost by using ChatGPT, instead of hiring freelancers through online platforms. On the supply side, it is important to acknowledge that freelancers can also leverage ChatGPT to improve their productivity, such as generating initial drafts and checking grammar efficiently. However, such productivity gains are insufficient to offset the decline in job opportunities. As a result, the translation & localization OLM experiences a significant net displacement effect.

## Effects on Web Development Freelancers

Our second analysis tests the effect of ChatGPT on web developers, using comparable workers in the construction design OLM as the control group. After matching with a caliper value of $4.6 \times 10^{-5}$, we obtained data for 3,139 workers. Table 5 reports the balance test results before and after the matching. Table 6 reports the DiD estimation results. The results of DiD analysis without matching are presented in Table A-2 of the Online Appendix A, demonstrating consistent findings.

In contrast to the results for translation & localization workers, we find opposite effects. Specifically, we find a 6.4% (= $e^{0.062} - 1$) increase in transaction volume for web developers after ChatGPT became available, as is suggested by the estimated coefficient of $ChatGPT_{it}$ in column (1). Furthermore, the estimated coefficient of $ChatGPT_{it}$ in column (2), corresponding to the dependent variable $\log(Fjobearn_{it})$, is positive and statistically significant, with a magnitude of nearly 66.5% (= $e^{0.510} - 1$).

These results indicate that ChatGPT is unlikely to automate the process of web development, but acts as an assistant to improve a web developer's productivity. Because web development jobs involve a variety of different tasks and require careful planning, ChatGPT alone cannot complete such jobs. On the demand side, employers still need to hire freelancers for such complex projects. While some relatively simple web development jobs may be independently handled by employers with the help of ChatGPT, the resulting reduction in market demand is limited. On the supply side, freelancers can effectively utilize ChatGPT to enhance their programming efficiency, such as generating basic code modules or debugging code. These productivity gains substantially improve freelancers' work efficiency, outweighing the negative effects stemming from reduced demand. Thus, the launch of ChatGPT eventually exerts a net productivity effect on the web development OLM.

## Parallel Trend Assumption

The parallel trend assumption and the no-anticipation assumption are key to the validity of DiD analysis. To provide empirical support, we conduct a lead-and-lag test, by estimating the following

relative-time model:

$$Y_{it} = \beta_0 + \sum_{\sigma=-6}^{\sigma=9} \beta_\sigma \times RelTime_\sigma \times Treat_i + \beta_2 \times X_{it} + \eta_i + \tau_t + \epsilon_{it} \quad (2)$$

In Equation (2), $RelTime_\sigma$ is a binary variable, which represents the relative month $\sigma$ to the release month of ChatGPT. $Treat_i$ is 1 if worker $i$ is in the treated occupation, and is 0 otherwise. We omit the first month prior to the release of ChatGPT, which serves as the baseline period. The set of coefficients $\beta_\sigma$ indicates whether different trends between workers in treated and control OLMs exist before ChatGPT's release ($\sigma < 0$) and how the estimated effects evolve over time afterward ($\sigma \geq 0$).

We report the results in Table 7 for translation & localization and web development jobs. We also visualize these coefficient estimations in Figure B-4 and Figure B-5 in the Online Appendix B. For all dependent variables and both markets, we find that the estimated coefficients $\beta_\sigma$ are insignificant before ChatGPT's release, which is consistent with our identification assumptions. The effects on $\log(Fjobnum_{it})$ and $\log(Fjobearn_{it})$ become significantly negative or positive after ChatGPT's release in each analysis. Interestingly, we find that the negative effect of ChatGPT on the transaction volume of translation jobs seems to strengthen over time, especially after March 2023. This finding shows that employers may need some time to assess the feasibility of substituting ChatGPT for translators. In contrast, the positive impact of ChatGPT on web development emerged early and remained relatively stable. One possible explanation is that web developers had prior exposure to AI tools, such as GitHub Copilot, which facilitated the immediate and effective integration of ChatGPT into their workflows.

## Extended Timeframe Analysis

To provide a richer understanding of ChatGPT and subsequent AI advancements on online labor markets in a longer time window, we extended the time frame to January 2025 and re-estimated the effects of ChatGPT's launch using the same PSM-DiD methodology and lead-and-lag test. Estimation results of DiD analysis, presented in Table 8, remain statistically significant and qualitatively consistent with our main analysis, confirming that our findings are not confined to a specific time window. Estimates from the relative-time model are shown in Table A-3 of Online Appendix A. The results show that the displacement effect in the translation & localization OLM persists and intensifies after December 2023. In contrast, the productivity effect in the web development OLM gradually diminishes and appears to taper off after December 2023. These patterns coincide with OpenAI's series of major product announcements in December 2023, including the release of OpenAI o1, Canvas, ChatGPT Pro, and other tools that significantly enhanced GPT's capabilities in writing and programming. This finding supports our interpretation that the effects are driven by advancements in AI.

## Robustness Checks

### Generalized Synthetic Control Method

In the previous analysis, we mainly used propensity score matching to improve sample balance. Here, we adopted an alternative matching method, namely the generalized synthetic control method (GSC), to create weighted control units and compare them with treatment units for each dependent variable [48]. The GSC method can also account for time-varying factors in the matching process and hence further enhance the empirical rigor. Specifically, we follow the prior literature [48] and employ a non-parametric bootstrap procedure to estimate average treatment effects. Our findings reveal substantial declines in translation & localization OLM in terms of log(*Fjobnum*) (−0.055***) and log(*Fjobearn*) (−0.168***). Web development OLM experiences significant increases in terms of log(*Fjobnum*) (0.038***) and log(*Fjobearn*) (0.208***). These estimation results are consistent with our main analyses in Table 4 and Table 6.

We then applied the two-one-sided *t* (TOST) test for equivalence tests, as shown in Figure B-6 and Figure B-7 in the Online Appendix B. The test results show that the average prediction error (gray dotted line) for all pretreatment periods lies within the equivalence range (red dotted line) for each dependent variable. This outcome confirms that there is no pretreatment trend before ChatGPT's release in both translation & localization and web development OLMs, compared to the construction design OLM, supporting the validity of our causal inference.

**Poisson Regression Estimates for Fjobnum**

In our main analysis, we log-transformed *Fjobnum* for the DiD estimations. Given the count nature of this variable, we conducted a robustness check by using Poisson regression. The results, shown in Table A-4 in Online Appendix A, remain consistent with our main findings. ChatGPT significantly reduced the number of jobs in the translation & localization OLM, while increasing job numbers for freelancers in the web development OLM.

**Control for Market-specific Time Trend**

A potential identification threat to the DiD strategy is an unobserved time-varying factor that affects different groups differently. To control for market-specific trends, we introduced categorical variables representing each distinct market. Then, we incorporated interactions between these categorical variables and time variables to account for varying trends across different markets. Table A-5 in Online Appendix A reports the estimated results. Again, we find that the translation & localization OLM experiences significant displacement effects, as evidenced by notable decreases in log(*Fjobnum*) and log(*Fjobearn*). On the other hand, the web development OLM demonstrates substantial productivity effects, with significant increases in all dependent variables. These findings are well aligned with our main analyses.

**Alternative Criterion of Active Freelancers**

Previously, we defined active freelancers as those who had accepted at least one job on the platform prior to the release of ChatGPT. To address potential concerns about the sensitivity of our results to this definition, we applied an alternative criterion in this section. Specifically, we redefined active freelancers as those who had completed at least one job before May 1, 2022, and re-estimated our models accordingly. The results, presented in Table A-6 of the Online Appendix A, are also consistent with our main analysis.

**Alternative Control Group and RDiT Analysis**

Despite the inclusion of various robustness checks, concerns may still arise regarding the validity of the control group used in the DiD analysis. To address this, we conducted an additional analysis using the Network & System Administration OLM as an alternative control group. The results remain consistent with our main findings, as detailed in Appendix C.

Furthermore, to complement the DiD approach, we employed a Regression Discontinuity in Time (RDiT) design, which leverages the timing of the ChatGPT release and does not rely on control group selection (see Appendix D) [23, 30]. As anticipated, the construction design OLM, serving as the control group, exhibits no statistically significant effect. In contrast, the translation & localization OLM shows pronounced displacement effects, whereas the web development OLM reveals significant productivity effects. Overall, the RDiT analyses provide further corroboration for the robustness of our main results.

The RDiT framework, by design, emphasizes short-term responses and is therefore more sensitive to transitory factors, such as seasonal slowdowns in December. In markets where the relevant tasks are not yet fully replicable by AI, or where freelancers and clients require time to adapt to new technologies, the immediate impacts captured by RDiT may be muted. In contrast, the DiD specification, which excludes November and December, is better suited to capture structural adjustments unfolding over a longer horizon. Therefore, we use RDiT as a robustness check and continue to use DiD as the main identification strategy. A comparative discussion of the differences between the RDiT and DiD estimates is presented in Appendix D.

# The Inflection Point Conjecture

## Inflection Point of AI and Jobs

Why does the exactly same AI innovation have opposite effects on freelancers of the two labor markets in our empirical study? We believe this seemingly simple question has a deeper answer worth a careful examination. To this end, we develop a microeconomic model to reveal the underlying economic mechanisms driving the empirical findings. Consider a Cournot competition model with $n$ workers each providing the same service with the same marginal cost of producing one unit of service.[5] Let the marginal cost be $(1-a)c$ where $c > 0$ and $a \in [0,1]$. We interpret $a$ as the percentage of tasks that can be successfully completed by AI during the production of the service. So $c$ represents a worker's marginal cost without using any AI assistance. Market demand for the service is determined by $p = S(a) - b\sum q_i$ where $p$ is the price, $q_i$ is the quantity of services provided by worker $i$, and $S(a)$ represents the market potential, which is decreasing in $a$. For potential employers who are more AI literate, AI is more reliable and competent in their focal jobs, which makes them more inclined to substitute AI for human labor. As AI improves, i.e., an increase in $a$, more potential employers fall into that category, thereby reducing the market potential. Moreover, $S(a)$ is likely concave because technology adoption often accelerates as the technology matures. There are several possible mechanisms. First, as AI performance increases, more employers will use it, which creates more word-of-mouth recommendations, hence more adoptions. Second, there is a positive externality from more employers using AI due to the dissemination of know-how and best practices. Third, innovative businesses may develop specialized software to facilitate the use of AI to aid specific occupations, as AI becomes increasingly powerful for that type of job. Fourth, our assumption of a concave function for $S(a)$ is also supported by the point–application–system framework proposed by Agrawal et al. [3], which outlines three stages of AI deployment and integration in industry. In the initial point solution stage, AI enhances performance on narrow, well-defined tasks, primarily augmenting human labor with minimal reductions in overall market potential. As AI advances to an application solution, it becomes more embedded in workflows by bundling multiple prediction tasks, leading to modest labor displacement. Ultimately, once AI reaches the system solution stage, AI enables end-to-end transformation of work processes, leading to significant labor displacement. This transition implies that as AI capability increases, the marginal reduction in market potential becomes steeper, justifying the concavity of $S(a)$.

[5] Providing a service is equivalent to a job or a project as clarified earlier.

The concavity of $S(a)$ is not uniform across all jobs but is shaped by the nature of the tasks involved. In particular, jobs with stable workflows, even if technically complex, can be more readily automated once AI-enabled tools are integrated into the entire process, accelerating the decline in market potential. Jobs with low stakes also tend to experience faster adoption, as employers are more willing to rely on AI despite imperfect accuracy. By contrast, jobs with unstable workflows or high stakes typically require ongoing human oversight, even at advanced stages of AI capability, leading to a more gradual reduction in $S(a)$.

Under the boundary conditions of $|S'(0)| < c$ and $|S'(1)| > c$, the equation $S'(a^*) + c = 0$ has a unique solution $a^* \in (0, 1)$.[6] The detailed derivation process is provided in Online Appendix E. We refer to $a^*$ as the AI inflection point for the focal market, as is justified by the following proposition.

**Proposition 1 (Inflection Point)** *Each worker enjoys higher job volume and more profit whenever AI level increases, up to the point of* $a^*$*, after which further increase in AI level reduces both job volume*

[6] $|S'(0)| < c$: When $a$ equals 0, AI has no capability to complete any tasks, so human workers perform the entire service. In this scenario, since AI has no influence, the demand potential reduction rate should be smaller than the marginal cost ($c$) to ensure that the initial impact of introducing AI is limited and does not drastically reduce the market potential.
$|S'(0)| > c$: When $a$ equals 1, AI can complete 100% of the tasks, meaning human involvement becomes unnecessary. In this case, the market potential drops drastically because there is no incentive to hire human workers anymore. Thus, the demand reduction rate ($|S'(1)|$) should exceed the marginal cost reduction to reflect the scenario where AI could substitute human labor, causing the market potential to decrease substantially.

*and profit. Moreover, a worker's revenue also decreases in AI level after it crosses the inflection point (i.e., $a > a^*$).*

We refer to the above model prediction as the *inflection point conjecture*. Intuitively, $a$ represents the degree to which AI replaces human labor. As $a$ increases, the potential market demand for human labor decreases, leading to fewer job opportunities and reduced profits for human workers at the micro level. This illustrates the negative impact of AI on labor demand, namely the displacement effect. Conversely, on the supply side, AI enhances worker productivity by assisting with various tasks, reducing the time and effort needed to complete them. This efficiency gain lowers costs and enables workers to take on more jobs, boosting both their transaction numbers and profits. An equilibrium between these contrasting effects is reached when $a$ is equal to $a^*$, which is exactly the inflection point of a market.

Clearly, different markets have different inflection points, which should be determined by the inherent characteristics of each job and AI capabilities. Before the inflection point, the marginal impact of any AI improvement is dominated by AI's productivity effect. But after the inflection point, the marginal impact of any AI improvement is dominated by AI's displacement effect.

We believe the contrasting effects of ChatGPT on the two OLMs analyzed thus far can be explained by the inflection point conjecture. For the translation & localization OLM, the jobs involved language-based tasks, such as sentence rewriting, grammar correction, and terminology substitution. These tasks are typically embedded in predictable and stable workflows, with relatively low consequences for minor errors, which reduces the perceived risk of bypassing human freelancers. As a result, $S(a)$ decreases rapidly, and clients become increasingly inclined to complete jobs independently using AI, causing a notable reduction in market potential. While AI can help freelancers complete tasks more

efficiently — thus lowering their effective cost $(1 - a)c$ — this gain is often outweighed by declining demand and growing competition. Consequently, the market surpasses its inflection point, at which displacement effects begin to dominate when AI capabilities advance.

In contrast, the web development OLM involves tasks such as system architecture, code integration, debugging, and customization based on client needs. These typically require ongoing coordination, and often evolve dynamically throughout a project. Unlike translation, web development workflows are generally less standardized and less stable, and errors tend to have higher stakes, such as causing system failures or compromising security. These attributes slow the changes of $S(a)$ with the rise of $a$, as clients continue to rely on human expertise to manage complexity and ensure reliability. On the supply side, freelancers can utilize generative AI to accelerate tasks involving code checking, function retrieval, and basic module referencing, thereby reducing their execution cost $(1 - a)c$ without experiencing a major drop in demand. This results in the scenario before the inflection point, where AI serves as a productivity-enhancing tool.

## Analyses of Additional Job Markets

The translation & localization OLM and the web development OLM may have represented two extreme cases of the effect of ChatGPT on freelancers. To understand the full spectrum, we obtained transaction data from other OLMs to further examine the impact of ChatGPT in those markets. Given the extensive expertise required to properly define each market, it is challenging to accurately select "specialties" based solely on our knowledge. Therefore, for these additional analyses, we primarily relied on the platform's classification system to determine which markets to select and which "specialties" on the platform should be included for each market.

As shown in Figure 1, while market definitions at the "category" or "specialty" level are either too broad or too narrow, market definition at the "subcategory" level seems well aligned with our intuition of what constitutes a market in practice. Thus, excluding the "category" that the construction design market belongs to (i.e., the control group), we chose "subcategory" from the remaining eleven "categories", and collected data from the "specialties" within the "subcategory" for each market. This approach not only ensures the breadth of our examination (i.e., covering every "category" on the platform) but also helps maintain a reasonable identification of markets (i.e., clustering "specialties" under each "subcategory" as an OLM). More specifically, we require: 1) the number of workers in each specialty is not too large (i.e., within the platform's maximum retrievable capacity) so that we can access all workers and their accepted jobs; and 2) the number of workers in each OLM is not too small (around 1,000) so that we have enough statistical power.[7]

To estimate the effect of ChatGPT on these eleven additional OLMs, we adopted the same sampling and identification strategies as in our main analyses. We summarize the coefficient estimates of the DiD estimators for different dependent variables in each of those eleven OLMs in Table 9. The detailed estimation results for each OLM are listed in Online Appendix F. For ease of comparison, we also list the corresponding estimates for the two OLMs in our main analyses. Overall, results show that OLMs closely tied to text generation, such as writing and translation jobs, experience substantial displacement effects. OLMs involved with code generation, such as web, mobile, and software development, network administration, and data science jobs, exhibit significant productivity effects. Furthermore, OLMs requiring high-level creativity, professional skills, and human interaction also show significant productivity effects.

[7] For each broad "category", we selected one "subcategory" that satisfied both criteria in our analysis. As a robustness check, we relaxed the selection criteria and included an additional "subcategory" from each "category". The key estimation results of these additional analyses, summarized in Table F-12 of the Online Appendix F, are consistent with our main findings.

To enhance the robustness of our findings, we further conducted analysis by integrating the specifications outlined in Section Robustness Checks — specifically restricting the sample to freelancers active prior to May 1, 2022, and incorporating market-specific time trends. Using this approach, we re-examined the results reported in Table 9. The outcomes, shown in Table 10, remain fully consistent with our main findings. We also performed RDiT analyses for all additional markets and found qualitatively similar results (see Appendix D). The ChatGPT effects on certain markets weaken or turn insignificant under RDiT estimations, likely because the RDiT approach focuses on only a narrow time window, and certain impacts may require more time to materialize. Taken together, these analyses provide additional support for the inflection point conjecture.

## Advance from ChatGPT 3.5 to 4.0

Considering the upgrade of ChatGPT from 3.5 to 4.0 as an additional leap in AI capabilities, we further investigate the evolving impact of LLM on freelancers. Compared with ChatGPT 3.5, ChatGPT 4.0 exhibits notable improvements in four key areas [39]. First, it extends from text-only processing to multimodality, accepting both text and image inputs. Second, it delivers substantially higher performance on professional and academic benchmarks, indicating stronger reasoning ability and enhanced coding proficiency. Third, it supports much longer contextual windows, which facilitates more coherent and nuanced responses in extended interactions and complex scenarios. Finally, it achieves greater factual accuracy and reduces hallucinations, benefiting from larger model scale, more diverse training data, and improved techniques such as reinforcement learning from human feedback (RLHF) [39].

Specifically, our study period includes two consecutive AI advancements, one on November 30, 2022 (i.e., release time for ChatGPT 3.5), and the other on March 14, 2023 (i.e., release time for ChatGPT 4.0). We used the following two-way fixed-effect DiD model to estimate the effects of both ChatGPT

3.5 and 4.0.

$$Y_{it} = \beta_0 + \beta_{1,1} \times ChatGPT3.5_{it} + \beta_{1,2} \times ChatGPT4.0_{it} + \beta_2 \times X_{it} + \eta_i + \tau_t + \epsilon_{it} \quad (3)$$

In Equation (3), the binary variable $ChatGPT3.5_{it}$ equals 1 if worker $i$ mainly belongs to the treated market and the transaction activities under investigation occurred after the release of ChatGPT 3.5; otherwise, the binary variable equals 0. Similarly, $ChatGPT4.0_{it}$ equals 1 if worker $i$ mainly belongs to the treated market and the transaction activities under investigation occurred after the release of ChatGPT 4.0; otherwise, it equals 0. We estimate the effects of ChatGPT 3.5 and 4.0 both for the two OLMs in our main analyses and for the eleven additional OLMs.

We summarize the estimated coefficients of variables of interest in Table 11 and report the detailed estimation results for each OLM in Online Appendix F. Specifically, $\beta_{1,1}$ measures the effect of AI advancement from the pre-3.5 period to the introduction of ChatGPT 3.5, while $\beta_{1,2}$ reflects the additional effect of the transition from ChatGPT 3.5 to 4.0. The sum, $\beta_{1,1} + \beta_{1,2}$, represents the total effects of AI progress from the pre-3.5 stage to version 4.0. Our results show that, compared to the initial introduction of version 3.5, the incremental improvement to version 4.0 had a more modest impact and, in many cases, adversely affected OLMs. This pattern suggests that as ChatGPT advances further, it begins to cross the inflection point for many OLMs, where the displacement effect outweighs productivity gains. The total effects on job volume and earnings from pre-3.5 to version 4.0 remain largely aligned with the effect types identified in our main analysis of the ChatGPT 3.5 release. Notably, for the information security & compliance OLM, the total effect is statistically insignificant, suggesting that the productivity gains observed after the release of ChatGPT 3.5 may have been offset by the displacement effect induced by the launch of ChatGPT 4.0.

To more easily compare the effects of the two shocks across different OLMs, we visualize the effects of ChatGPT on each OLM in Figure 3 where each dot corresponds to an OLM, and the horizontal and vertical coordinates represent the first-shock effect (i.e., advancement from previous version to ChatGPT 3.5) and the second-shock effect (i.e., advancement from ChatGPT 3.5 to 4.0), respectively. We also use the size of a dot to represent the level of statistical significance. There are three scenarios following two leaps in AI capabilities: 1) continued net productivity effect: AI remains below the inflection point after both upgrades. 2) net productivity effect to net displacement effect: AI does not reach the inflection point after the first upgrade but surpasses it following the second upgrade. This transitional pattern demonstrates how successive upgrades can fundamentally alter the balance between productivity gains and labor substitution due to AI advancement, validating the existence of market-specific inflection points. 3) continued net displacement effect: AI has already surpassed the inflection point with the first upgrade and continues to exceed it after the second upgrade. Accordingly, these three scenarios correspond to dots in the first, the fourth, and the third quadrant, respectively.

The absence of any dot in the second quadrant is consistent with our inflection point conjecture: once AI crosses a critical threshold of capability, its displacement effect tends to become self-reinforcing and difficult to reverse. In the broader AI discourse, concerns about over-automation and the misalignment between AI capabilities and actual task requirements have sparked debates on de-automation, re-skilling, and even re-humanizing certain roles (e.g., in education, counseling, or creative industries) [6, 13]. However, both our theoretical model and empirical findings suggest that such reversals are unlikely to occur. Rather than anticipating a return to pre-AI labor structures, workers and institutions must proactively adapt their career strategies in response to this structural shift.

We further categorize these OLMs into five clusters based on their job content and effects, as summarized in Table F-26 of the Online Appendix F. Specifically, writing-related jobs (e.g., translation) are highly susceptible to displacement effects, as they primarily involve text generation—a domain where ChatGPT has demonstrated strong capability. With the improvements of ChatGPT 4.0, especially its enhanced comprehension of nuanced prompts and ability to produce more contextually appropriate outputs, writing jobs such as copywriting, business writing, and research reports are likely to face even greater displacement risks as revision cycles shorten and content quality rises. Consulting-related jobs (e.g., legal consulting) rely on specialized expertise and decision support. While these jobs initially benefited from productivity gains, ChatGPT 4.0's stronger contextual understanding and domain knowledge, illustrated by its top 10% performance on the simulated Bar Exam, suggest that some consulting jobs may gradually transition toward displacement as AI continues to expand its knowledge base [39]. Programming-related jobs (e.g., web development) initially experienced strong productivity enhancements, as ChatGPT assists developers with debugging and generating code modules. However, ChatGPT 4.0's improved coding performance indicates that certain programming jobs once requiring freelance support may increasingly be automated, raising the prospect of future displacement effects. Operational jobs (e.g., project management) involve domain-specific knowledge, contextual understanding, and human interaction, which limit the potential for displacement and lead primarily to productivity gains. Creativity-related jobs (e.g., product design) require originality and open-ended ideation. While ChatGPT 4.0 can support early-stage idea generation, it cannot yet fully replicate humans' creativity, making displacement unlikely in the short term.

## Effects of ChatGPT on Fulfilled Demand

In previous analyses, we have explored how the relative position between AI intelligence and inflection points affects freelancers. In this section, we further investigate the effects of ChatGPT on the total (fulfilled) demand.

To this end, it is critical to obtain complete transaction data from each OLM to calculate the volume of fulfilled job postings at the market level. This proxy is used as an indirect signal of the demand-side trends. Unfortunately, due to compliance with platform policies, we cannot access all job posting records in the translation & localization and web development sectors. These sectors contain highly popular "specialties" attracting such a large number of freelancers that it exceeds the platform's data retrieval limits. Consequently, we have focused this analysis on the additional eleven markets where full transaction data is available.

To ensure an adequate sample size, we extended our previous time window to make it span from January 2022 to October 2023. We used the following two-way fixed-effect DiD model for identification where the unit of analysis is at the market-week level.

$$\log\left(Postnum_{jt}\right) = \beta_0 + \beta_1 \times ChatGPT_{jt} + \eta_j + \tau_t + \epsilon_{jt} \tag{4}$$

In Equation (4), $j$ and $t$ index market and week, respectively. The dependent variable $\log(Postnum_{jt})$ measures the log-transformed number of the volume of fulfilled job postings in market $j$ during week $t$. The explanatory variable of interest is the binary variable $ChatGPT_{jt}$ which equals 1 if market $j$ is the treated market and week $t$ is after the release of ChatGPT, and 0 otherwise. $\eta_j$ captures the market fixed effect, while $\tau_t$ captures the time fixed effect.

We summarize the estimated coefficients for the eleven OLMs in Table 12. The results indicate that the release of ChatGPT has significant effects on the fulfilled demand in these markets. Specifically, in markets where we previously found a net displacement effect of AI, we observe a decrease in the total number of fulfilled jobs. In contrast, in markets where we previously found a net productivity effect of AI, we observe an increase in the total number of fulfilled jobs.

# Heterogeneous Analysis

## Freelancer Location

ChatGPT, developed by an American AI research organization OpenAI, profoundly shocked society and markets in the United States. As a result, the number of American users ranks first among all regions. This analysis hence investigates whether American freelancers are more or less affected by ChatGPT compared to those in other regions. We introduced $US_i$ as the moderator variable, defined as 1 if freelancer $i$ resides in the United States, and 0 otherwise. We included the binary variable $US_i$ and its interaction with $ChatGPT_{it}$ into the regression. Note the variable $US_i$ itself is absorbed by the freelancer fixed effect.

Table 13 reports the estimations for both translation & localization and web development markets. The results indicate that the location of the freelancer does not significantly affect ChatGPT's impact in the translation & localization markets. This result makes sense because, in our model, the displacement effect mainly stems from reduced demand for freelancers. Thus, the location of the freelancer, a supply-side factor related to whether a freelancer can easily leverage ChatGPT for productivity enhancement, should not matter much in markets where the displacement effect dominates. In contrast, US-based web developers experience greater productivity effects. This finding aligns with our model, where such effects primarily stem from worker productivity enhancements facilitated by AI. Specifically, US-based web developers are likely to have better access to and greater familiarity with ChatGPT, which, in turn, amplifies the productivity effect they experience.

## Freelancer Experience

Experienced workers have been identified as more aware of market dynamics and potential threats [19], potentially reacting differently to ChatGPT's release compared to their less experienced counterparts. This varied response could further lead to shifts in the composition of suppliers in certain markets, hence motivating us to examine AI impact heterogeneity over the freelancer experience.

Specifically, we calculated the total number of focal jobs each freelancer accepted before the release of ChatGPT and defined the moderator variable $Experienced_i$ as 1 if the number of focal jobs accepted by freelancer $i$ before the release is above the median, and 0 otherwise. We then included the interactions of this binary variable $Experienced_i$ with relevant variables in the regression. Note the variable $Experienced_i$ itself is absorbed by the freelancer fixed effect.

Table 14 reports the estimation results, indicating two contrasting effects of freelancers' experience on supply and demand sides. On the supply side, AI empowers inexperienced workers by lowering the skill threshold and enabling them to take on a wider range of tasks. In contrast, the productivity gains for experienced workers are relatively limited, as they already possess the necessary skills to perform such tasks efficiently. Therefore, from a supply-side perspective, inexperienced workers are more likely to benefit from AI. On the demand side, the introduction of AI reduces the demand for simple jobs, as the related tasks become increasingly automated. This trend disproportionately affects inexperienced workers who tend to accept such jobs. Consequently, from a demand-side perspective, inexperienced workers are more likely to be disadvantaged.

In translation & localization OLM, the productivity boost for inexperienced translators outweighs the automation of simple translation jobs, resulting in a comparatively smaller negative impact on inexperienced translators compared with experienced translators. However, in web development OLM, the productivity boost for inexperienced developers is balanced by the automation of simple projects, leading to no significant heterogeneous effects between inexperienced and experienced workers.

# Conclusions

The ongoing debate concerning the interplay between AI and human labor has been characterized by two contrasting views, emphasizing either the displacement effect or the productivity effect. On one hand, there are concerns about skill displacement that human labor might be replaced by AI. On the other hand, there are observations that AI could augment human productivity and even create enough new job opportunities. The recent rise of LLMs marks a pivotal change in the landscape of AI, significantly altering how we live and work, which has also sparked global apprehension again about potential technological unemployment. Different from previous AI tools, LLMs like ChatGPT have demonstrated remarkable performance in language-related tasks. A wide range of markets have been exposed to this popular tool. How LLMs substitute or complement human labor needs more empirical investigations. This study constitutes an early effort in this important endeavor.

This paper contributes both empirically and theoretically to our understanding of AI's implications on workers, especially on freelancers. On the empirical side, this research is among the first to document two opposite scenarios of the AI–freelancer relationship, the occurrence of which depends on the interplay between AI and the task components of an OLM. The primary example of the first scenario is the OLM of translation & localization, which experienced significant displacement effects, while the OLM of web development benefited from substantial productivity effects. Extending the analysis to a broader range of OLMs, our study underscores the depth and reach of this wave of AI innovations. By leveraging the introduction of ChatGPT 3.5 and 4.0 as two successive shocks, we adopt a dynamic perspective to capture how AI's impact evolves over time. The results reveal a transition scenario, where productivity effects gradually shift into displacement, and indicate that markets that have crossed the inflection point are unlikely to return to a productivity-enhancing phase. The moderation analyses further highlight that the same AI shock can generate heterogeneous outcomes across different freelancer groups.

On the theoretical front, we proposed the inflection point conjecture, emphasizing that the dichotomy exposed by our empirical analyses is not static. There is no fundamental difference between OLMs suffering from the relentless encroachment of AI and OLMs benefiting from exactly the same tools. What differs across different OLMs is their respective inflection points. Specifically, before AI performance crosses the inflection point associated with an OLM, freelancers always benefit from an improvement in AI performance. However, after AI performance crosses the inflection point, freelancers become worse off whenever AI performance further improves. By adopting a technology-agnostic framework, our model moves beyond task-based perspectives that may become outdated as AI's frontier advances. Our micro-level model also provides a more nuanced understanding of the short-term dynamics of AI adoption and aligns closely with our empirical findings. Importantly, while our model introduces the notion of an inflection point at which advancement in AI capability begins to hurt rather than augment human labor, it does not imply that all markets will eventually be replaced by AI. The actual trajectory of AI adoption is influenced by a range of complex factors—including technical feasibility, regulatory and ethical constraints, and the value of uniquely human skills—which may prevent certain markets from ever reaching the inflection point $a^*$. Our framework is thus best understood as an interpretable tool for identifying when and why displacement effects emerge, rather than a deterministic forecast of universal job replacement.

We believe our findings have important practical implications for the future of work. In particular, our study highlights the evolving role of AI in benefiting or hurting workers' job prospects as technology progresses. Workers, therefore, should be cognizant of not just the current role of AI in their jobs but its future trajectory. For example, workers in labor markets already in the substitution phase of their relation with AI should actively seek other careers because AI encroachment will only deepen as AI improves. On the other hand, workers currently benefiting from an AI-induced productivity boost cannot be complacent either. These fortunate workers should stay vigilant of any sign of AI crossing the inflection point so that they can plan ahead. Our model also offers insights into welfare implications. As AI capabilities improve, employers may achieve more efficient output from freelancers at the same cost or lower labor expenses by replacing human workers with AI tools. For freelancers, surplus increases as AI increases up to the inflection point, after which their surplus decreases as AI further increases.

Our paper has several limitations which we hope future research can address. First of all, our empirical analyses are based on data collected from one freelance platform. While we believe the mechanisms we identify are likely applicable to other platforms, the magnitude and timing of AI's effects may vary depending on platform-specific factors. Given the importance of the topic, there is an urgent need for more studies using data from other freelance platforms. Second, our findings are based on freelance markets which may not generalize to full-time jobs. Full-time jobs often face stricter legal and privacy constraints on AI use and typically benefit from contractual protections that reduce the risk of displacement. Further analyses using other data sources are needed to generalize to full-time employment settings. Third, the fulfilled demand used in our analysis does not fully represent overall market demand. Examining trends in total job postings before and after the release of ChatGPT would offer a more comprehensive view of demand-side dynamics, if such data can be accessed. Fourth, future research could also incorporate more granular project-level data to yield deeper insights into how AI reshapes the composition and dynamics of work within freelance platforms. Fifth, we examined the effect of only one technology leap in AI, albeit a major one. As the current wave of AI innovations unfolds, there are plenty of opportunities to examine the effects of other major innovations. Finally, the inflection point conjecture, despite its structural insight, falls short of predicting anything quantitative, which limits its practical value. Future research could address this limitation by developing empirical approaches to estimate the percentage of tasks within a job that can be successfully completed by AI ($a$) and constructing more practical proxies for the inflection point ($a^*$) across different jobs.

# Tables

**Table 1: Definitions of Key Variables**

| Variables | Definitions |
|---|---|
| $Fjobearn_{it}$ | The total earnings of focal jobs accepted in month *t* by worker *i* |
| $Fjobnum_{it}$ | The number of focal jobs accepted in month *t* by worker *i* |
| $Fjobprice_{it}$ | The average price per focal job accepted in month *t* by worker *i* |
| $Fhourprice_{it}$ | The average hourly rate of focal jobs accepted in month *t* by worker *i* |
| $Fjobrating_{it}$ | The average rating of focal jobs accepted in month *t* by worker *i* |
| $Tenure_{it}$ | The number of months since worker *i*'s registration up to month *t* |

**Table 2: Descriptive Statistics of Key Variables**

| Measure | Count | Mean | Std. dev. | Min | Max |
|---|---|---|---|---|---|
| **Construction Design OLM** | | | | | |
| $Fjobearn_{it}$ | 86688 | 181.289 | 1244.317 | 0.000 | 66754.383 |
| $Fjobnum_{it}$ | 86688 | 0.308 | 0.980 | 0.000 | 43.000 |
| $Fjobprice_{it}$ | 14312 | 705.461 | 2261.041 | 1.000 | 66754.383 |
| $Fhourprice_{it}$ | 5442 | 26.101 | 17.354 | 3.000 | 180.000 |
| $Fjobrating_{it}$ | 8445 | 4.873 | 0.430 | 1.000 | 5.000 |
| $Tenure_{it}$ | 86688 | 32.626 | 30.318 | 0.000 | 230.000 |
| **Translation & Localization OLM** | | | | | |
| $Fjobearn_{it}$ | 91744 | 111.641 | 747.553 | 0.000 | 46785.879 |
| $Fjobnum_{it}$ | 91744 | 0.433 | 1.473 | 0.000 | 124.000 |
| $Fjobprice_{it}$ | 17449 | 319.918 | 1085.959 | 0.650 | 43915.000 |
| $Fhourprice_{it}$ | 4793 | 21.613 | 15.023 | 3.000 | 500.000 |
| $Fjobrating_{it}$ | 9726 | 4.926 | 0.329 | 1.000 | 5.000 |
| $Tenure_{it}$ | 91744 | 38.929 | 33.022 | 0.000 | 197.000 |
| **Web Development OLM** | | | | | |
| $Fjobearn_{it}$ | 172448 | 646.368 | 4439.452 | 0.000 | 294652.500 |
| $Fjobnum_{it}$ | 172448 | 0.399 | 1.093 | 0.000 | 45.000 |
| $Fjobprice_{it}$ | 35398 | 2226.121 | 7684.738 | 1.000 | 294652.500 |
| $Fhourprice_{it}$ | 20019 | 29.062 | 22.495 | 3.000 | 500.000 |
| $Fjobrating_{it}$ | 25458 | 4.851 | 0.491 | 1.000 | 5.000 |
| $Tenure_{it}$ | 172448 | 45.559 | 38.651 | 0.000 | 280.000 |

Note: If worker *i* does not accept any focal jobs in month *t*, $Fjobprice_{it}$, $Fhourprice_{it}$, and $Fjobrating_{it}$ would be recorded as a null value, and $Fjobratio_{it}$ would be recorded as zero.

**Table 3: Propensity Score Matching: Balance Test Between Treated (Translation & Localization) and Control (Construction Design) Groups**

| | Prematching | | | | Postmatching | | | |
|---|---|---|---|---|---|---|---|---|
| | Mean treated | Mean control | t-test $p > \|t\|$ | Std. diff. | Mean treated | Mean control | t-test $p > \|t\|$ | Std. diff. |
| Accumulative *Fjobnum* | 3.153 | 2.659 | 0.000 | 0.412 | 3.019 | 2.993 | 0.504 | 0.022 |
| Accumulative *Experience* | 3.628 | 3.327 | 0.000 | 0.344 | 3.518 | 3.556 | 0.174 | −0.044 |
| Average *Fjobprice* | 5.423 | 5.994 | 0.000 | −0.462 | 5.620 | 5.630 | 0.787 | −0.009 |
| Average *Fhourprice* | 2.760 | 2.893 | 0.000 | −0.241 | 2.803 | 2.785 | 0.312 | 0.033 |
| Average *Fjobrating* | 4.913 | 4.857 | 0.000 | 0.202 | 4.908 | 4.895 | 0.116 | 0.045 |

Note: This research utilizes the log transformation of *Fjobnum*, *Experience*, *Fjobprice*, and *Fhourprice*. Subsequent variable processing for PSM follows the same methodology.

**Table 4: Effect of ChatGPT on Translation & Localization Jobs**

| | Variables | |
|---|---|---|
| | (1) log(*Fjobnum*) | (2) log(*Fjobearn*) |
| ChatGPT | −0.094*** | −0.353*** |
| | (0.014) | (0.072) |
| Observations | 36416 | 36416 |
| *N* | 2276 | 2276 |
| Adj. $R^2$ | 0.469 | 0.344 |

Note: (1) *$p$<0.1, **$p$<0.05, ***$p$<0.01; (2) Clustered standard errors are in the parentheses; (3) We control for time fixed effect, worker fixed effect and worker tenure. Unless otherwise noted, the same specifications are applied in the subsequent tables.

**Table 5: Propensity Score Matching: Balance Test Between Treated (Web Development) and Control (Construction Design) Groups**

| | Prematching | | | | Postmatching | | | |
|---|---|---|---|---|---|---|---|---|
| | Mean treated | Mean control | t-test $p > \|t\|$ | Std. diff. | Mean treated | Mean control | t-test $p > \|t\|$ | Std. diff. |
| Accumulative *Fjobnum* | 2.777 | 2.659 | 0.000 | 0.106 | 2.844 | 2.834 | 0.747 | 0.009 |
| Accumulative *Experience* | 3.516 | 3.328 | 0.000 | 0.202 | 3.477 | 3.463 | 0.575 | 0.015 |
| Average *Fjobprice* | 6.940 | 5.991 | 0.000 | 0.645 | 6.409 | 6.394 | 0.632 | 0.010 |
| Average *Fhourprice* | 2.975 | 2.892 | 0.000 | 0.144 | 2.905 | 2.914 | 0.554 | −0.016 |
| Average *Fjobrating* | 4.852 | 4.857 | 0.377 | −0.020 | 4.853 | 4.861 | 0.284 | −0.028 |

**Table 6: Effect of ChatGPT on Web Development Jobs**

| | Variables | |
|---|---|---|
| | (1) log(*Fjobnum*) | (2) log(*Fjobearn*) |
| ChatGPT | 0.062*** | 0.510*** |
| | (0.011) | (0.065) |
| Observations | 50224 | 50224 |
| *N* | 3139 | 3139 |
| Adj. $R^2$ | 0.357 | 0.269 |

Note: (1) *$p$<0.1, **$p$<0.05, ***$p$<0.01; (2) Clustered standard errors are in the parentheses.

**Table 7: Relative-time Model: Effects of ChatGPT on Translation & Localization and Web Development Jobs**

| | Translation & Localization Jobs | | Web Development Jobs | |
|---|---|---|---|---|
| | (1) log(*Fjobnum*) | (2) log(*Fjobearn*) | (3) log(*Fjobnum*) | (4) log(*Fjobearn*) |
| $RelTime_{t-6}$ | −0.036 | −0.155 | −0.012 | −0.098 |
| | (0.027) | (0.150) | (0.022) | (0.145) |
| $RelTime_{t-5}$ | 0.005 | 0.250 | −0.021 | −0.117 |
| | (0.028) | (0.158) | (0.023) | (0.145) |
| $RelTime_{t-4}$ | −0.011 | 0.069 | 0.001 | −0.046 |
| | (0.025) | (0.145) | (0.021) | (0.145) |
| $RelTime_{t-3}$ | 0.013 | 0.089 | 0.027 | 0.143 |
| | (0.025) | (0.140) | (0.022) | (0.145) |
| $RelTime_{t-2}$ | 0.005 | 0.165 | 0.013 | 0.102 |
| | (0.023) | (0.137) | (0.020) | (0.134) |
| $RelTime_t$ | −0.077*** | −0.251* | 0.070*** | 0.554*** |
| | (0.026) | (0.149) | (0.022) | (0.143) |
| $RelTime_{t+1}$ | −0.079*** | −0.196 | 0.055*** | 0.432*** |
| | (0.024) | (0.135) | (0.021) | (0.134) |
| $RelTime_{t+2}$ | −0.067*** | −0.170 | 0.066*** | 0.504*** |
| | (0.026) | (0.138) | (0.022) | (0.143) |
| $RelTime_{t+3}$ | −0.110*** | −0.352** | 0.044** | 0.401*** |
| | (0.025) | (0.143) | (0.022) | (0.143) |
| $RelTime_{t+4}$ | −0.096*** | −0.255* | 0.060*** | 0.540*** |
| | (0.025) | (0.148) | (0.021) | (0.138) |
| $RelTime_{t+5}$ | −0.095*** | −0.301** | 0.047** | 0.336** |
| | (0.027) | (0.147) | (0.023) | (0.150) |
| $RelTime_{t+6}$ | −0.096*** | −0.276** | 0.068*** | 0.566*** |
| | (0.025) | (0.137) | (0.021) | (0.137) |
| $RelTime_{t+7}$ | −0.105*** | −0.364*** | 0.062*** | 0.559*** |
| | (0.025) | (0.140) | (0.023) | (0.143) |
| $RelTime_{t+8}$ | −0.134*** * | −0.372*** | 0.074*** | 0.632*** |
| | (0.025) | (0.135) | (0.023) | (0.146) |
| $RelTime_{t+9}$ | −0.123*** | −0.296** | 0.084*** | 0.542*** |
| | (0.025) | (0.135) | (0.022) | (0.142) |
| Observations | 36416 | 36416 | 50224 | 50224 |
| *N* | 2276 | 2276 | 3139 | 3139 |
| Adj. $R^2$ | 0.469 | 0.344 | 0.357 | 0.269 |

Note: (1) *$p$<0.1, **$p$<0.05, ***$p$<0.01; (2) Clustered standard errors are in the parentheses.

**Table 8: Additional Analysis: Extended Time Frame from May 2022 to January 2025**

| | Translation & Localization Jobs | | Web Development Jobs | |
|---|---|---|---|---|
| | (1) | (2) | (3) | (4) |
| | log(*Fjobnum*) | log(*Fjobearn*) | log(*Fjobnum*) | log(*Fjobearn*) |
| ChatGPT | −0.103*** | −0.409*** | 0.045*** | 0.322*** |
| | (0.015) | (0.073) | (0.013) | (0.079) |
| Observations | 53289 | 53289 | 66092 | 66092 |
| Adjusted $R^2$ | 0.436 | 0.340 | 0.372 | 0.284 |

Note: (1) *$p$<0.1, **$p$<0.05, ***$p$<0.01; (2) Clustered standard errors are in the parentheses.

**Table 9: Effects of ChatGPT on Different OLMs**

| Category | Specific OLM | log(*Fjobnum*) | log(*Fjobearn*) |
|---|---|---|---|
| Translation | Translation & Localization | −0.094*** | −0.353*** |
| Web, Mobile & Software Development | Web Development | 0.062*** | 0.510*** |
| Writing | Professional & Business Writing | −0.079*** | −0.390*** |
| Translation | Language Tutoring & Interpretation | −0.071** | −0.159 |
| IT & Networking | Information Security & Compliance | 0.055** | 0.292* |
| Accounting & Consulting | Financial Planning | 0.074*** | 0.425*** |
| Web, Mobile & Software Development | Game Design & Development | 0.091*** | 0.563*** |
| Customer Service | Community Management & Tagging | 0.092*** | 0.673*** |
| Legal | Corporate & Contract Law | 0.122*** | 0.515*** |
| Admin Support | Project Management | 0.100*** | 0.694*** |
| Data Science & Analytics | Data Mining & Management | 0.153*** | 0.895*** |
| Sales & Marketing | Marketing, PR & Brand Strategy | 0.191*** | 1.251*** |
| Design & Creative | Photography | 0.214*** | 1.018*** |

Note: (1) *$p$<0.1, **$p$<0.05, ***$p$<0.01.

**Table 10: Effects of ChatGPT on Different OLMs with More Stringent Empirical Strategy**

| Category | Specific OLM | log(*Fjobnum*) | log(*Fjobearn*) |
|---|---|---|---|
| Translation | Translation & Localization | −0.055*** | −0.287*** |
| Web, Mobile & Software Development | Web Development | 0.056*** | 0.359*** |
| Writing | Professional & Business Writing | −0.042** | −0.253** |
| Translation | Language Tutoring & Interpretation | −0.079* | −0.230 |
| IT & Networking | Information Security & Compliance | 0.109*** | 0.575*** |
| Accounting & Consulting | Financial Planning | 0.122*** | 0.662*** |
| Web, Mobile & Software Development | Game Design & Development | 0.065** | 0.334* |
| Customer Service | Community Management & Tagging | 0.110** | 0.664* |
| Legal | Corporate & Contract Law | 0.095** | 0.557** |
| Admin Support | Project Management | 0.057* | 0.451** |
| Data Science & Analytics | Data Mining & Management | 0.097*** | 0.543*** |
| Sales & Marketing | Marketing, PR & Brand Strategy | 0.085*** | 0.559*** |
| Design & Creative | Photography | 0.098*** | 0.480*** |

Note: (1) *$p$<0.1, **$p$<0.05, ***$p$<0.01; (2) We control for time fixed effect, worker fixed effect, worker tenure, and market-specific time trend.

**Table 11: Effects of ChatGPT 3.5 and 4.0 on Different OLMs**

| Specific OLM | log(*Fjobnum*) | | | log(*Fjobearn*) | | |
|---|---|---|---|---|---|---|
| | ChatGPT3.5 ($\beta_{1,1}$) | ChatGPT4.0 ($\beta_{1,2}$) | Total Effect $\beta_{1,1}$+$\beta_{1,2}$ | ChatGPT3.5 ($\beta_{1,1}$) | ChatGPT4.0 ($\beta_{1,2}$) | Total Effect $\beta_{1,1}$+$\beta_{1,2}$ |
| Translation & Localization | −0.074*** | −0.025* | −0.099*** | −0.293*** | −0.075 | −0.368*** |
| Web Development | 0.061*** | 0.001 | 0.062*** | 0.496*** | 0.017 | 0.513*** |
| Professional & Business Writing | −0.045** | −0.043*** | −0.088*** | −0.236** | −0.193* | −0.429*** |
| Language Tutoring & Interpretation | −0.037 | −0.044 | −0.080** | −0.065 | −0.118 | −0.183 |
| Information Security & Compliance | 0.106*** | −0.064* | 0.042 | 0.472** | −0.225 | 0.247 |
| Financial Planning | 0.100*** | −0.032* | 0.067*** | 0.491*** | −0.082 | 0.409*** |
| Game Design & Development | 0.070*** | 0.026 | 0.096*** | 0.431*** | 0.164 | 0.595*** |
| Community Management & Tagging | 0.110** | −0.023 | 0.087** | 0.726** | −0.066 | 0.659** |
| Corporate & Contract Law | 0.126*** | −0.005 | 0.121*** | 0.561** | −0.058 | 0.504** |
| Project Management | 0.078*** | 0.027 | 0.105*** | 0.559*** | 0.169 | 0.728*** |
| Data Mining & Management | 0.124*** | 0.036 | 0.160*** | 0.702*** | 0.242* | 0.944*** |
| Marketing, PR & Brand Strategy | 0.131*** | 0.074*** | 0.206*** | 0.918*** | 0.416*** | 1.334*** |
| Photography | 0.139*** | 0.094*** | 0.233*** | 0.664*** | 0.442*** | 1.106*** |

Note: (1) *$p$<0.1, **$p$<0.05, ***$p$<0.01; (2) $\beta_{1,1}$ denotes effect of AI advancement from previous version to ChatGPT 3.5; (3) $\beta_{1,2}$ denotes effect of AI advancement from ChatGPT 3.5 to 4.0.

**Table 12: Effects of ChatGPT on Fulfilled Demand for Different OLMs**

| OLM | Dominating Effects | log(*Postnum*) |
|---|---|---|
| Professional & Business Writing | Displacement | −0.276*** |
| Language Tutoring & Interpretation | Displacement | −0.196*** |
| Information Security & Compliance | Productivity | 0.197*** |
| Financial Planning | Productivity | 0.155*** |
| Game Design & Development | Productivity | 0.158*** |
| Community Management & Tagging | Productivity | 0.133*** |
| Corporate & Contract Law | Productivity | 0.156*** |
| Project Management | Productivity | 0.257*** |
| Data Mining & Management | Productivity | 0.206*** |
| Marketing, PR & Brand Strategy | Productivity | 0.057*** |
| Photography | Productivity | 0.251*** |

Note: (1) *$p$<0.1, **$p$<0.05, ***$p$<0.01.

**Table 13: Heterogeneous Analyses Over Freelancers' Location**

| | Translation & Localization Jobs | | Web Development Jobs | |
|---|---|---|---|---|
| | (1) | (2) | (3) | (4) |
| | log(*Fjobnum*) | log(*Fjobearn*) | log(*Fjobnum*) | log(*Fjobearn*) |
| US×ChatGPT | 0.015 | −0.020 | 0.096** | 0.493* |
| | (0.054) | (0.254) | (0.047) | (0.286) |
| ChatGPT | −0.095*** | −0.347*** | 0.056*** | 0.487*** |
| | (0.014) | (0.075) | (0.012) | (0.067) |
| US×After | 0.007 | 0.168 | −0.007 | 0.076 |
| | (0.031) | (0.192) | (0.028) | (0.166) |
| Observations | 36416 | 36416 | 50224 | 50224 |
| N | 2276 | 2276 | 3139 | 3139 |
| Adj. $R^2$ | 0.469 | 0.344 | 0.357 | 0.269 |

Note: (1) *$p$<0.1, **$p$<0.05, ***$p$<0.01; (2) Clustered standard errors are in the parentheses.

**Table 14: Heterogeneous Analyses Over Freelancers' Experience**

| | Translation & Localization Jobs | | Web Development Jobs | |
|---|---|---|---|---|
| | (1) | (2) | (3) | (4) |
| | log(*Fjobnum*) | log(*Fjobearn*) | log(*Fjobnum*) | log(*Fjobearn*) |
| Experienced×ChatGPT | −0.102*** | −0.241* | 0.009 | −0.042 |
| | (0.028) | (0.144) | (0.023) | (0.132) |
| ChatGPT | −0.044*** | −0.235*** | 0.058*** | 0.530*** |
| | (0.015) | (0.081) | (0.012) | (0.076) |
| Experienced×After | 0.008 | 0.006 | −0.024 | −0.052 |
| | (0.020) | (0.111) | (0.017) | (0.097) |
| Observations | 36416 | 36416 | 50224 | 50224 |
| *N* | 2276 | 2276 | 3139 | 3139 |
| Adj. $R^2$ | 0.470 | 0.344 | 0.357 | 0.269 |

Note: (1) *$p$<0.1, **$p$<0.05, ***$p$<0.01; (2) Clustered standard errors are in the parentheses.

# Figures

## Figure 1: Classification System on the Online Freelance Platform

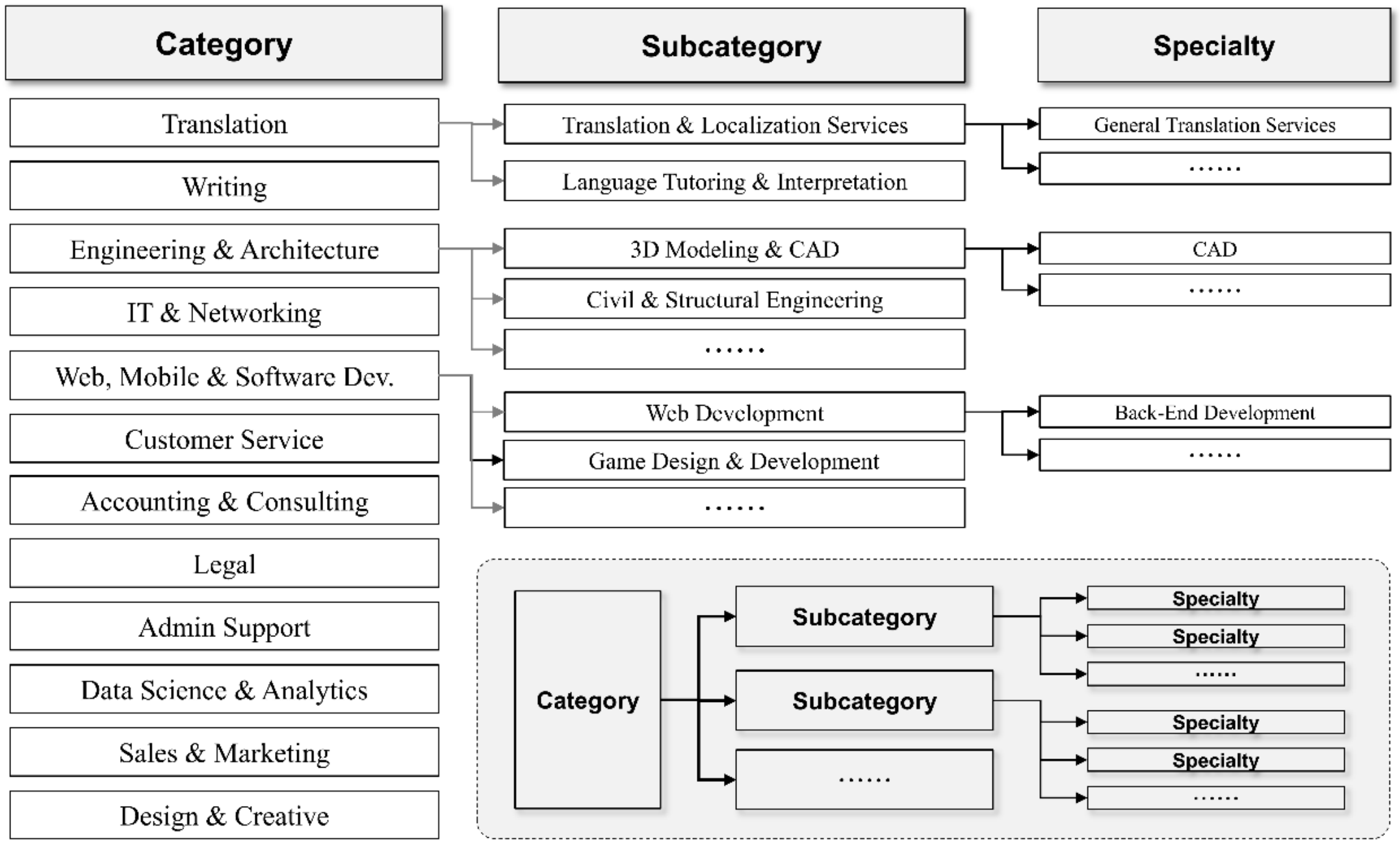

## Figure 2: Empirical Framework: Effects of ChatGPT on OLMs

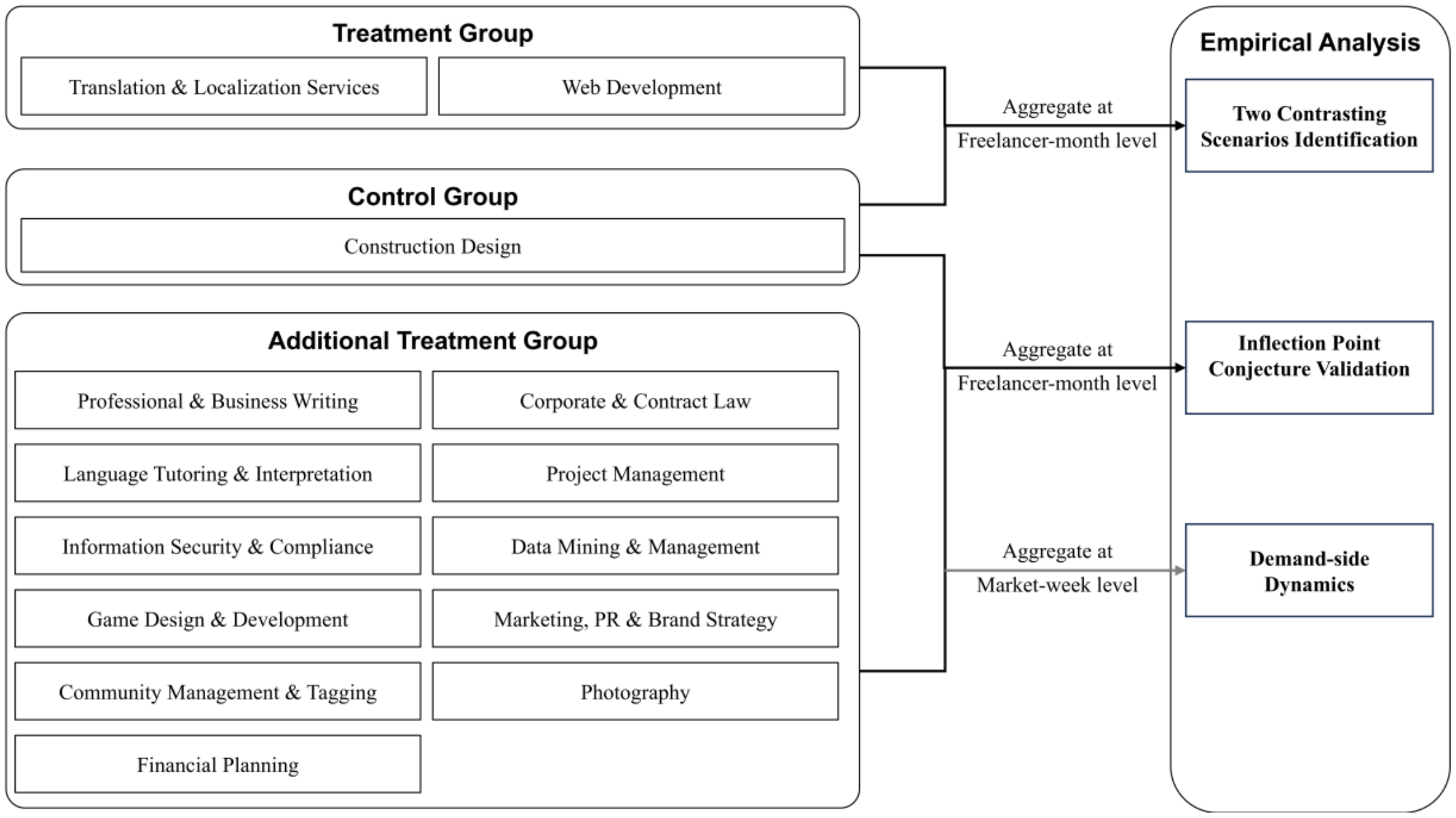

**Figure 3: Effects of ChatGPT 3.5 and 4.0 across Different OLMs on a Coordinate Plane**

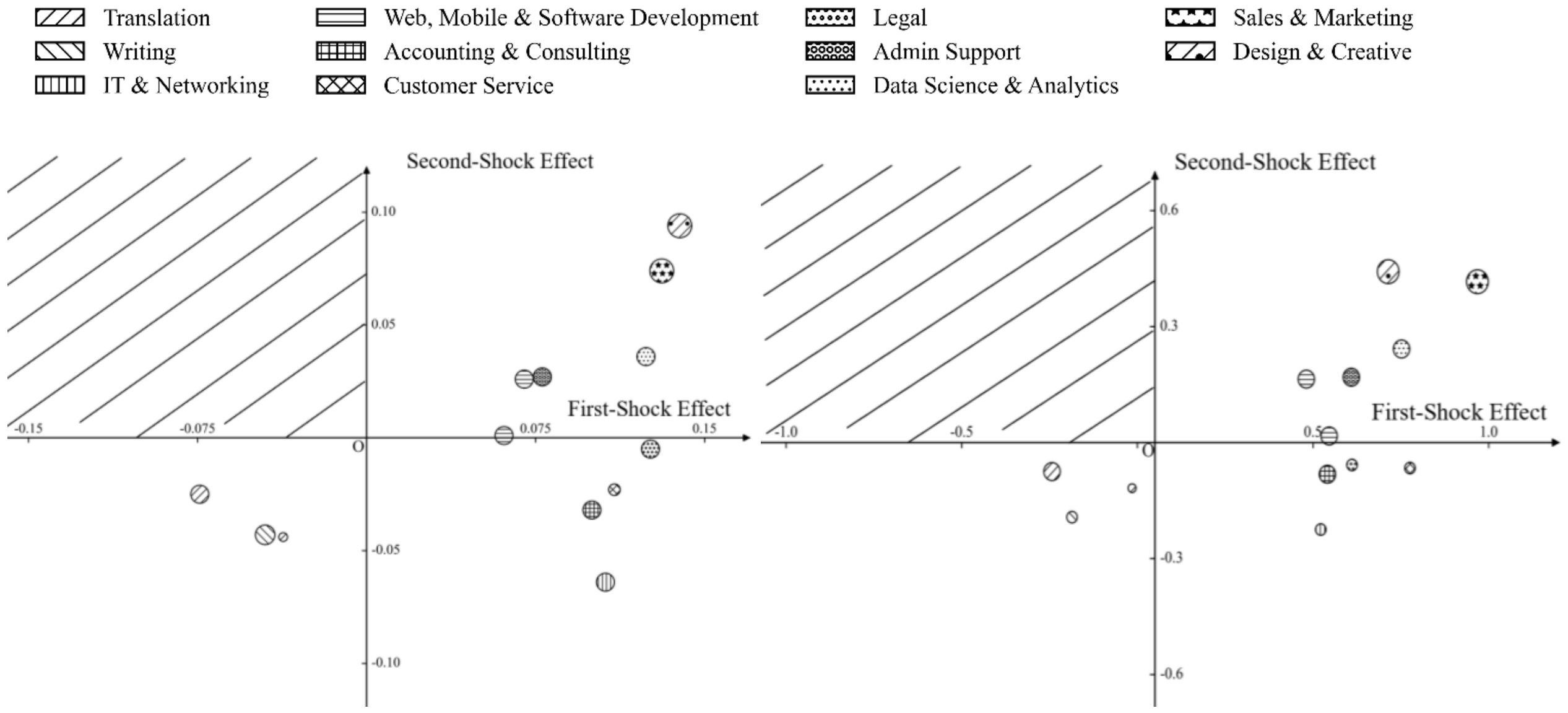

(a) log(*Fjobnum*)

(b) log(*Fjobearn*)